\documentclass[MSNbibl,nameyear,seceqn,dvips]{arxstspdf}
\usepackage{multirow}
\usepackage{algorithm}
\usepackage{graphicx}
\usepackage{flushend}
\usepackage{stfloats}

\volume{27}
\issue{4}
\pubyear{2012}
\firstpage{519}
\lastpage{537}
\doi{10.1214/12-STS391} 

\makeatletter
\newtheorem{theorem}{Theorem}[section]

\newtheorem{proposition}{Proposition}[section]
\newtheorem{lemma}{Lemma}[section]
\newproclaim{remark}{Remark}[section]
\newproclaim{example}{Example}[section]
\newproclaim{assumption}{Assumption}[section]
\newproclaim{definition}{Definition}[section]

\renewcommand{\citep}[1]{\citeauthor{#1}, \citeyear{#1}}
\newcommand{\eqref}[1]{(\ref{#1})}

\def\X{\mathcal{X}}
\def\truep{p^*}
\def\div{\|}
\def\reals{{\mathbb R}}
\def\P{{\mathbb P}}
\def\argmin{\operatorname{arg\,min}}
\def\argmax{\operatorname{arg\,max}}
\let\hat\widehat
\let\tilde\widetilde
\let\hat\widehat

\def\except{\backslash}
\def\npn{\operatorname{\mathit{NPN}}}
\def\i{{(i)}}
\def\cG{\mathcal{G}}

\makeatother

\begin{document}
\begin{frontmatter}

\title{Sparse Nonparametric Graphical Models}

\runtitle{Sparse nonparametric graphical models}

\begin{aug}
\author[a]{\fnms{John} \snm{Lafferty}\corref{}\ead[label=e1]{lafferty@uchicago.edu}},
\author[b]{\fnms{Han} \snm{Liu}\ead[label=e2]{hanliu@princeton.edu}}
\and
\author[c]{\fnms{Larry} \snm{Wasserman}\ead[label=e3]{larry@stat.cmu.edu}}

\runauthor{J. Lafferty, H. Liu and L. Wasserman}


\address[a]{John Lafferty is Professor, Department of Statistics and Department of Computer Science,
University of Chicago, 5734 S. University Avenue, Chicago, Illinois 60637, USA \printead{e1}.}

\address[b]{Han Liu is Assistant Professor, Department of Operations Research and Financial Engineering, Princeton University,
Princeton, New Jersey 08544, USA \printead{e2}.}

\address[c]{Larry Wasserman is Professor, Department of Statistics
and Machine Learning Department,Carnegie Mellon University, Pittsburgh Pennsylvania 15213, USA \printead{e3}.}
\end{aug}

%
\begin{abstract}
We present some nonparametric methods for graphical
modeling. In the discrete case, where the data are binary or drawn
from a finite alphabet, Markov random fields are already essentially
nonparametric, since the cliques can take only a finite number of
values. Continuous data are different. The Gaussian graphical model
is the standard parametric model for continuous data, but it makes
distributional assumptions that are often unrealistic. We discuss
two approaches to building more flexible graphical models. One allows
arbitrary graphs and a nonparametric extension of the Gaussian; the
other uses kernel density estimation and restricts the graphs to
trees and forests. Examples of both methods are presented.
We also discuss possible future research directions for nonparametric
graphical modeling.
\end{abstract}

%
\vspace*{-5pt}
\end{frontmatter}

\section{Introduction}

This paper presents two methods for constructing nonparametric
graphical models for continuous data. In the discrete case, where the
data are binary or drawn from a finite alphabet, Markov random fields
or log-linear models are already essentially nonparametric, since the
cliques can take only a finite number of values. Continuous data are
different. The Gaussian graphical model is the standard parametric
model for continuous data, but it makes distributional assumptions\vadjust{\goodbreak} that
are typically unrealistic. Yet few practical alternatives to the
Gaussian graphical model exist, particularly for high-dimen\-sional~data.
We discuss two approaches to building more flexible graphical models
that exploit sparsity. These two approaches are at different extremes
in the array of choices available. One allows arbitrary graphs, but
makes a distributional restriction through the use of copulas; this is
a semiparametric extension of the Gaussian. The other approach uses
kernel density estimation and restricts the graphs to trees and
forests; in this case the model is fully nonparametric, at the expense
of structural restrictions. We describe two-step estimation methods
for both approaches. We also outline some statistical theory for the
methods, and compare them in some examples. This article is in part a
digest of two recent research articles where these methods first
appeared, \citet{npn:09} and \citet{fde:11}.

\begin{figure*}
\begin{tabular}{@{}lll@{}}
\hline
& \textbf{Nonparanormal} & \textbf{Forest densities} \\
\hline
Univariate marginals & nonparametric & nonparametric \\
Bivariate marginals & determined by Gaussian copula & nonparametric \\
Graph & unrestricted & acyclic \\
\hline
\end{tabular}
\caption{Comparison of properties of the nonparanormal and
forest-structured densities.} \label{fig::npn-fd-compare}
\end{figure*}

The methods we present here are relatively simple, and many more
possibilities remain for nonparametric graphical modeling. But as we
hope to demonstrate, a~little nonparametricity can go a~long way.

\section{Two Families of Nonparametric Graphical Models}

The graph of a random vector is a useful way of exploring the
underlying distribution. If $X=(X_1,\ldots,\break X_d)$ is a random vector
with distribution $P$, then the undirected graph $G=(V,E)$
corresponding to $P$ consists of a vertex set $V$ and an edge set $E$
where $V$ has $d$ elements, one for each variable $X_i$. The edge
between $(i,j)$ is excluded from $E$ if and only if $X_i$ is
independent of $X_j$, given the other variables $X_{\except\{i,j\}}
\equiv(X_s\dvtx1\leq s \leq d, s\neq i,j)$, written
%
\begin{equation}\label{eq::ci}
X_i \amalg X_j | X_{\except\{i,j\}}.
\end{equation}

The general form for a (strictly positive) probability density encoded
by an undirected graph $G$ is
%
\begin{equation}\label{eq:npgm}
p(x) = \frac{1}{Z(f)} \exp\biggl( \sum_{C\in\mathrm{Cliques}(G)}
f_C(x_C) \biggr),
\end{equation}
where the sum is over all cliques, or fully connected subsets of
vertices of the graph. In general, this is what we mean by a
\textit{nonparametric graphical
model}. It is the graphical model analog of the general
nonparametric regression model. Model~\eqref{eq:npgm} has two main
ingredients, the graph $G$ and the functions $\{f_C\}$. However,
without further assumptions, it is much too general to be practical.
The main difficulty in working with such a model is the normalizing
constant $Z(f)$, which cannot, in general, be efficiently computed or
approximated.

In the spirit of nonparametric estimation, we can seek to impose
structure on either the graph or the functions $f_C$ in order to get a
flexible and useful family of models. One approach parallels the ideas
behind sparse additive models for regression. Specifically, we replace
the random variable $X=(X_1,\ldots,\break X_d)$ by the transformed random
variable $f(X) = (f_1(X_1), \ldots, f_d(X_d) )$, and assume that
$f(X)$ is multivariate Gaussian. This results in a nonparametric
extension of the Normal that we call the \textit{nonparanormal}
distribution. The nonparanormal depends on the univariate functions
$\{f_j\}$, and a mean $\mu$ and covariance matrix $\Sigma$, all of
which are to be estimated from data. While the resulting family of
distributions is much richer than the standard parametric Normal (the
paranormal), the independence relations among the variables are still
encoded in the precision matrix $\Omega= \Sigma^{-1}$, as we show
below.

The second approach is to force the graphical structure to be a tree or
forest, where each\vadjust{\goodbreak} pair of vertices is connected by at most one path.
Thus, we relax the distributional assumption of normality, but we
restrict the allowed family of undirected graphs. The complexity of the
model is then regulated by selecting the edges to include, using cross
validation.

Figure~\ref{fig::npn-fd-compare} summarizes the tradeoffs made by these
two families of models. The nonparanormal can be thought of as an
extension of additive models for regression to graphical modeling. This
requires estimating the univariate marginals; in the copula approach,
this is done by estimating the functions\break $f_j(x) = \mu_j + \sigma_j
\Phi^{-1}(F_j(x))$, where $F_j$ is the distribution function for
variable $X_j$. After estimating each $f_j$, we transform to (assumed)
jointly Normal via $Z = (f_1(X_1), \ldots, f_d(X_d))$ and then apply
methods for Gaussian graphical models to estimate the graph. In this
approach, the univariate marginals are fully nonparametric, and the
sparsity of the model is regulated through the inverse covariance
matrix, as for the graphical lasso, or ``glasso''
(Banerjee, El~Ghaoui and d'Aspremont, \citeyear{Banerjee:08};
Friedman, Hastie and Tibshirani, \citeyear{FHT:07}).\footnote{Throughout the paper we use the
term graphical lasso, or glasso, coined by \citet{FHT:07} to refer to
the solution obtained by $\ell_1$-regularized log-likelihood under the
Gaussian graphical model. This estimator goes back at least to
\citet{Yuan:Lin:07}, and an iterative lasso algorithm for doing the
optimization was first proposed by \citet{Banerjee:08}. In our
experiments we use the R packages \texttt{glasso} (Friedman, Hastie and Tibshirani, \citeyear{FHT:07}) and
\texttt{huge} to implement this algorithm.} The model is estimated in a
two-stage procedure; first the functions $f_j$ are estimated, and then
inverse covariance matrix $\Omega$ is estimated. The high-level
relationship between linear regression models, Gaussian graphical
models and their extensions to additive and high-dimensional models is
summarized in Figure~\ref{fig::compare}.

\begin{figure*}
\begin{tabular*}{\textwidth}{@{\extracolsep{\fill}}llll@{}}
\hline
\textbf{Assumptions} & \textbf{Dimension} & \textbf
{Regression}&\textbf{Graphical models} \\
\hline
Parametric & low & linear model & multivariate Normal
\\
& high & lasso & graphical lasso\\[6pt]
Nonparametric & low & additive model & nonparanormal\\
& high & sparse additive model & sparse nonparanormal\\
\hline
\end{tabular*}
\caption{Comparison of regression and graphical models. The
nonparanormal extends additive models to the graphical model
setting. Regularizing the inverse covariance leads to an extension
to high dimensions, which parallels sparse additive models for
regression.}\vspace*{-3pt}
\label{fig::compare}
\end{figure*}

In the forest graph approach, we restrict the graph to be acyclic, and
estimate the bivariate marginals $p(x_i,x_j)$ nonparametrically. In
light of equation \eqref{eq.treedensity}, this yields the full
nonparametric family of graphical models having acyclic graphs. Here
again, the estimation procedure is two-stage; first the marginals are
estimated, and then the graph is estimated. Sparsity is regulated
through the edges $(i,j)$ that are included in the forest.

Clearly these are just two tractable families within the very large
space of possible nonparametric graphical models specified by
equation~\eqref{eq:npgm}. Many interesting research possibilities
remain for novel nonparametric graphical models that make different
assumptions; we discuss some possibilities in a concluding section. We
now discuss details of these two model families, beginning with the
nonparanormal.

\section{The Nonparanormal}
\label{sec:npn}

We say that a random vector $X=(X_1,\ldots,X_d)^T$ has a {\em
nonparanormal} distribution and write
\[
X \sim\npn(\mu,\Sigma,f)
\]
in case\vspace*{1pt} there exist functions $\{f_j\}_{j=1}^d$ such that $Z \equiv
f(X) \sim N(\mu,\Sigma)$, where $f(X) = (f_1(X_1),\ldots, f_d(X_d))$.
When the $f_j$'s are monotone and differentiable, the joint
probability density function of $X$ is given by
%
\begin{eqnarray}\label{eq:npndensity}
p_X(x)& =& \frac{1}{(2\pi)^{d/2}|\Sigma|^{1/2}}\nonumber
\\[-1pt]
&&{}\cdot\exp\biggl\{-\frac{1}{2}
\bigl(f(x)-\mu\bigr)^{T}{\Sigma}^{-1} \bigl(f(x)-\mu\bigr)
\biggr\}
\\[-1pt]
&&{}\cdot\prod_{j=1}^d|f'_j(x_j)|,\nonumber
\end{eqnarray}
where the product term is a Jacobian.

Note that the density in \eqref{eq:npndensity} is not
identifiable---we could scale each function by a constant, and
scale the diagonal of $\Sigma$ in the same way, and not change the
density. To make the family identifiable we demand that $f_j$
preserves marginal means and variances.
\begin{eqnarray}\label{eq:identify}
\mu_{j} &=& \mathbb{E}(Z_{j}) = \mathbb{E}(X_j)\quad
\mbox{and}\nonumber
\\[-8pt]\\[-8pt]
\sigma^2_{j} &\equiv&\Sigma_{jj} = \operatorname{Var} (
Z_j )
= \operatorname{Var} ( X_j ).\nonumber
\end{eqnarray}
These conditions only depend on $\operatorname{diag}(\Sigma)$, but
not the
full covariance matrix.

Now, let $F_{j}(x)$ denote the marginal distribution function of
$X_{j}$. Since the component $f_j(X_j)$ is Gaussian, we have that
\begin{eqnarray*}
F_{j}(x) &=& \mathbb{P} (X_{j} \leq x ) \\
&=& \mathbb{P} \bigl(Z_j \leq f_j(x) \bigr)
= \Phi\biggl( \frac{f_j(x)-\mu_{j}}{\sigma_{j}} \biggr)
\end{eqnarray*}
which implies that
%
\begin{equation}\label{eq.marginalf}
f_j(x) = \mu_{j} + \sigma_{j} \Phi^{-1} (F_j(x) ).
\end{equation}
The form of the density in \eqref{eq:npndensity} implies that the
conditional independence graph of the nonparanormal is encoded in
$\Omega= \Sigma^{-1}$, as for the parametric Normal, since the density
factors with respect to the graph of $\Omega$, and therefore obeys the
global Markov property of the graph.

\begin{figure*}

\includegraphics{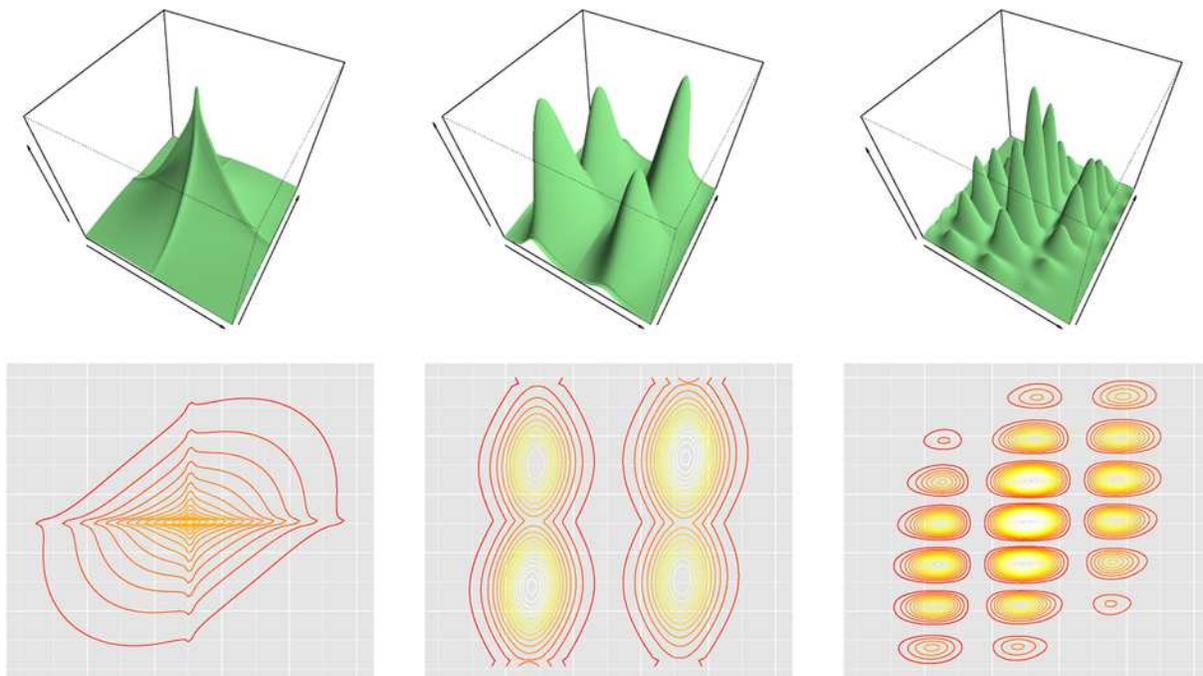}

%
\caption{Densities of three 2-dimensional nonparanormals. The left
plots have component functions of the form $f_\alpha(x) =\allowbreak
\operatorname{sign}(x)|x|^{\alpha}$, with $\alpha_1=0.9$ and
$\alpha_2=0.8$. The center plots have component functions
of the form $g_\alpha(x) = \allowbreak\lfloor x \rfloor+ 1/(1+\exp(-\alpha(x
- \lfloor x \rfloor- 1/2)))$ with
$\alpha_1=10$ and $\alpha_2=5$, where
$x - \lfloor x \rfloor$ is the fractional part.
The right plots have component functions
of the form $h_\alpha(x) = x + \sin(\alpha x)/\alpha$, with $\alpha_1=5$
and $\alpha_2=10$. In each case $\mu=(0,0)$ and $\Sigma=
\bigl({{1\enskip0.5}\atop{0.5\enskip1}}\bigr)$.\vspace*{2pt}}
\label{fig:densityex}
\end{figure*}

In fact, this is true for any choice of identification restrictions;
thus it is not necessary to estimate $\mu$ or $\sigma$ to estimate the
graph, as the following result shows.

\begin{lemma}
\label{lemma:h} Define
%
\begin{equation}\label{eq::h}
h_j(x) = \Phi^{-1}(F_j(x)),
\end{equation}
and let $\Lambda$ be the covariance matrix of $h(X)$. Then $X_j \amalg
X_k | X_{\except\{j,k\}}$ if and only if $\Lambda_{jk}^{-1} =0$.
\end{lemma}

\begin{pf}
We can rewrite the covariance matrix as
\[
\Sigma_{jk} = \operatorname{Cov}(Z_j,Z_k) = \sigma_{j}\sigma
_{k}\operatorname{Cov}(h_j(X_j),h_k(X_k)).
\]
Hence $\Sigma= D \Lambda D$ and
\[
\Sigma^{-1} = D^{-1} \Lambda^{-1} D^{-1},
\]
where $D$ is the diagonal matrix with $\operatorname{diag}(D) = \sigma
$. The
zero pattern of $\Lambda^{-1}$ is therefore identical to the zero
pattern of $\Sigma^{-1}$.
\end{pf}

Figure~\ref{fig:densityex} shows three examples of 2-dimensional
nonparanormal densities. The component functions are taken to be from\vadjust{\goodbreak}
three different families of monotonic functions---one using power
transforms, one using logistic transforms and another using sinusoids.
\begin{eqnarray*}
f_\alpha(x) & =& \operatorname{sign}(x) |x|^\alpha,\\
g_\alpha(x) & = &\lfloor x \rfloor+ \frac{1}{1+\exp\{-\alpha(x -
\lfloor x \rfloor- 1/2)\}}, \\
h_\alpha(x) & = & x + \frac{\sin(\alpha x)}{\alpha}.
\end{eqnarray*}
The covariance in each case is $\Sigma= \bigl({{1\enskip0.5}\atop
{0.5\enskip1}}\bigr)$, and the
mean is $\mu=(0,0)$. It can be seen how the concavity and number of
modes of the density can change with different nonlinearities. Clearly
the nonparanormal family is much richer than the Normal family.

The assumption that $f(X) = (f_1(X_1), \ldots, f_d(X_d))$ is Normal
leads to a semiparametric model where only one-dimensional functions
need to be estimated. But the monotonicity of the functions $f_j$,
which map onto $\reals$, enables computational tractability of the
nonparanormal. For more general functions $f$, the normalizing
constant for the density
\[
\label{eq:gpndensity}
p_X(x) \propto\exp\biggl\{-\frac{1}{2} \bigl(f(x)-\mu\bigr
)^{T}{\Sigma}^{-1}
\bigl(f(x)-\mu\bigr) \biggr\}
\]
cannot be computed in closed form.

\subsection{Connection to Copul\ae}

If $F_j$ is the distribution of $X_j$, then $U_j = F_j(X_j)$ is
uniformly distributed on $(0,1)$. Let $C$ denote the joint
distribution function of $U = (U_1,\ldots, U_d)$, and let $F$ denote
the distribution function of $X$. Then we have that
%
\begin{eqnarray}
&&F(x_1,\ldots, x_d)\nonumber
\\[-8pt]\\[-8pt]
&&\quad= \P(X_1\leq x_1, \ldots, X_d\leq x_d)\nonumber\\
&&\quad= \P\bigl(F_1(X_1)\nonumber
\\[-8pt]\\[-8pt]
&&\hphantom{= \P\bigl(}\quad\leq F_1(x_1), \ldots, F_d(X_d)\leq F_d(x_d)\bigr)\nonumber\\
&&\quad= \P\bigl(U_1\leq F_1(x_1), \ldots, U_d\leq F_d(x_d)\bigr)
\\
&&\quad= C(F_1(x_1),\ldots, F_d(x_d)).
\end{eqnarray}
This is known as Sklar's theorem (\citep{Sklar:59}), and $C$ is
called a \textit{copula}. If $c$ is the density function of~$C$, then
\begin{eqnarray}
&&p(x_1,\ldots, x_d) \nonumber
\\[-8pt]\\[-8pt]
&&\quad= c(F_1(x_1),\ldots, F_d(x_d)) \prod_{j=1}^d p(x_j),\nonumber
\end{eqnarray}
where $p(x_j)$ is the marginal density of $X_j$. For the nonparanormal
we have
\begin{eqnarray}
&&\quad F(x_1,\ldots,x_d)\nonumber\\[-8pt]\\[-8pt]
&&\qquad= \Phi_{\mu,\Sigma}( \Phi^{-1}(F_1(x_1)),\ldots,
\Phi^{-1}(F_d(x_d))),\nonumber
\end{eqnarray}
where $\Phi_{\mu,\Sigma}$ is the multivariate Gaussian cdf, and
$\Phi$
is the univariate standard Gaussian cdf.

The Gaussian copula is usually expressed in terms of the correlation
matrix, which is given by $R = \operatorname{diag}(\sigma)^{-1}
\Sigma
\operatorname{diag}(\sigma)^{-1}$. Note that the univariate mar\-
ginal density
for a Normal can be written as $p(x_j) = \frac{1}{\sigma_j} \phi(u_j)$
where $u_j = (x_j-\mu_j)/\sigma_j$. The
multivariate Normal density
can thus be expressed as
%
\begin{eqnarray}
&&p_{\mu,\Sigma}(x_1,\ldots, x_d) \nonumber
\\
&&\quad= \frac{1}{(2\pi)^{d/2}
|R|^{1/2} \prod_{j=1}^d \sigma_j }
\\
&&\qquad{}\cdot\exp\biggl(-\frac{1}{2} u^T R^{-1} u
\biggr)\nonumber\\
&&\quad= \frac{1}{|R|^{1/2}} \exp\biggl(-\frac{1}{2} u^T
(R^{-1}-I) u \biggr) \nonumber
\\[-8pt]\\[-8pt]
&&\qquad{}\cdot\prod_{j=1}^d \frac{\phi(u_j)}{\sigma_j}.\nonumber
\end{eqnarray}
Since the distribution $F_j$ of the $j$th variable satisfies $F_j(x_j)
= \Phi((x_j-\mu_j)/\sigma_j) = \Phi(u_j)$, we have that
$(X_j-\mu_j)/\sigma_j \stackrel{d}{=} \Phi^{-1}(F_j(X_j))$. The
Gaussian copula density is thus
%
\begin{eqnarray}
&&c(F_1(x_1),\ldots, F_d(x_d))\nonumber
\\
&&\quad= \frac{1}{|R|^{1/2}}
\exp
\biggl\{-\frac{1}{2} \Phi^{-1}(F(x))^T
\\
&&\qquad\hphantom{\frac{1}{|R|^{1/2}}
\exp
\biggl\{}{}\cdot(R^{-1}-I) \Phi^{-1}(F(x)) \biggr\},\nonumber
\end{eqnarray}
where
\[
\Phi^{-1}(F(x)) = (\Phi^{-1}(F_1(x_1)), \ldots,
\Phi^{-1}(F_d(x_d))).
\]
This is seen to be equivalent to
\eqref{eq:npndensity} using the chain rule and the identity
%
\begin{eqnarray}
(\Phi^{-1})'(\eta) =\frac{1}{\phi( \Phi^{-1}(\eta) )}.
\end{eqnarray}

\subsection{Estimation}
\label{sec:estimation}

Let $X^{(1)},\ldots, X^{(n)}$ be a sample of size $n$ where
$X^\i=(X^\i_{1},\ldots, X^\i_{d})^T \in\mathbb{R}^d$. We'll
design a
two-step estimation procedure where first the functions $f_j$ are\vadjust{\goodbreak}
estimated, and then the inverse covariance matrix $\Omega$ is
estimated, after transforming to approximately Normal.

In light of \eqref{eq::h} we define
\[
\hat{h}_j(x) = \Phi^{-1}(\tilde{F}_j(x)),
\]
where $\tilde{F}_j$ is an estimator of $F_j$. A natural candidate for
$\tilde{F}_j$ is the marginal empirical distribution function
\[
\hat{F}_j(t) \equiv\frac{1}{n}\sum_{i=1}^n\mathbf{1}_{ \{X^{(i)}_{j}
\leq t \}}.
\]
However, in this case $\hat h_j(x)$ blows up at the largest and
smallest values of $X_j^{(i)}$. For the high-dimensional setting where
$n$ is small relative to $d$, an attractive alternative is to use a
truncated or \textit{Winsorized\/}\footnote{After Charles P. Winsor, the
statistician whom John Tukey credited with his conversion from topology
to statistics (\citep{Tukey:VI:90}).} estimator,
%
\begin{equation}\label{eq:truncatedestimator}
\hspace*{8pt}\qquad\tilde{F}_j(x) =
\cases{
\delta_{n}, & \mbox{if $\hat{F}_j(x) < \delta_n$,} \cr
\hat{F}_j(x), & \mbox{if $\delta_n \leq\hat{F}_j(x) \leq1-\delta
_n$,} \cr
(1-\delta_n), & \mbox{if $\hat{F}_j(x) > 1- \delta_n$},
}\hspace*{-8pt}
\end{equation}
where $\delta_n$ is a truncation parameter. There is a bias--variance
tradeoff in choosing $\delta_n$; increasing $\delta_n$ increases the
bias while it decreases the variance.

Given this estimate of the distribution of variable $X_{j}$, we then
estimate the transformation function $f_j$ by
%
\begin{equation}\label{eq:keyestimator}
\tilde{f}_j(x) \equiv\hat{\mu}_{j}+\hat{\sigma}_{j}\tilde{h}_j(x),
\end{equation}
where
\[
\tilde{h}_j(x) = \Phi^{-1} (\tilde{F}_j(x) )
\]
and $\hat{\mu}_{j}$ and $\hat{\sigma}_{j}$ are the sample mean and
standard deviation.
\[
\hat{\mu}_{j} \equiv\frac{1}{n}\sum_{i=1}^{n}X^\i_{j}\quad\mbox
{and}\quad
\hat{\sigma}_{j}= \sqrt{\frac{1}{n}\sum_{i=1}^{n} \bigl(X^\i_{j} -
\hat{\mu}_{j} \bigr)^{2}}.
\]
Now, let $S_n(\tilde{f})$ be the sample covariance matrix of
$\tilde{f}(X^{(1)}),\ldots, \tilde{f}(X^{(n)})$; that is,
\begin{eqnarray}
\label{eq:covardef} S_n(\tilde{f}) &\equiv& \frac{1}{n}
\sum_{i=1}^n \bigl( \tilde{f}\bigl(X^\i\bigr) - \mu_n(\tilde
{f}) \bigr)\nonumber
\\[-9pt]\\[-9pt]
&&\hphantom{\frac{1}{n}
\sum_{i=1}^n}{}\cdot\bigl( \tilde
{f}\bigl(X^\i\bigr) - \mu_n(\tilde{f}) \bigr)^T,\nonumber\\[-2pt]
\mu_n(\tilde{f}) &\equiv& \frac{1}{n} \sum_{i=1}^n \tilde{f}\bigl
(X^\i\bigr).
\nonumber
\end{eqnarray}
We then estimate $\Omega$ using $S_n(\tilde f)$. For instance, the
maximum likelihood estimator is $\hat\Omega^{\mathrm{MLE}}_n =
S_n(\tilde
f)^{-1}$.\vadjust{\goodbreak}

The $\ell_1$-regularized estimator is
\begin{eqnarray}\label{eq:winsorized-est}
\hat\Omega_n &=& \operatorname{arg\,min}\limits_{\Omega}
\{\operatorname{tr} (\Omega S_n(\tilde f) ) \nonumber
\\[-9.5pt]\\[-9.5pt]
&&\hphantom{\operatorname{arg\,min}\limits_{\Omega}
\{}{}- \log|\Omega| +
\lambda\|\Omega\|_1 \},\nonumber
\end{eqnarray}
where $\lambda$ is a regularization parameter, and
$\|\Omega\|_1=\sum_{j=1}^{d}\sum_{k=1}^{d}|\Omega_{jk}|$. The
estimated graph is then $\hat{E}_n = \{ (j,k)\dvtx\hat\Omega
_{jk}\neq
0\}$.

Thus we use a two-step procedure to estimate the graph:\vspace*{-1pt}

\begin{longlist}
\item[(1)] Replace the observations, for each variable, by their
respective Normal scores, subject to a Winsorized truncation.
\item[(2)] Apply the graphical lasso to the transformed data to
estimate the undirected graph.\vspace*{-1pt}
\end{longlist}

The first step is noniterative and computationally efficient. The
truncation parameter $\delta_{n}$ is chosen to be
%
\begin{eqnarray}
\delta_{n} = \frac{1}{4n^{1/4}\sqrt{\pi\log n}}
\end{eqnarray}
and does not need to be tuned. As will be shown in Theorem
\ref{thm.keylemma}, such a choice makes the nonparanormal amenable to
theoretical analysis.\vspace*{-3pt}

\subsection{\texorpdfstring{Statistical Properties of $S_n(\tilde f)$}{Statistical Properties of $S n(f)$}}
\label{sec:npntheory}

The main technical result is an analysis of the covariance of the
Winsorized estimator above. In particular, we show that under
appropriate conditions,
\[
\max_{j,k} | S_n(\tilde{f})_{jk} - S_n(f)_{jk} | = O_{P} \Biggl(
\sqrt{\frac{\log d + \log^2 n}{n^{1/2}}} \Biggr),
\]
where $S_n(\tilde f)_{jk}$ denotes the $(j,k)$ entry of the matrix
$S_n(\tilde f)$. This result allows us to leverage the significant body
of theory on the graphical lasso (\citep{Rothman:08};
\citep{Ravikumar:Gauss:09}) which we apply in step two.\vspace*{-1pt}

\begin{theorem}\label{thm.keylemma}
Suppose that $d = n^\xi$, and let $\tilde{f}$ be the Winsorized
estimator defined in \eqref{eq:keyestimator} with $\delta_{n} =
\frac{1}{4n^{1/4}\sqrt{\pi\log n}}$. Define
\[
C(M,\xi) \equiv\frac{48}{\sqrt{\pi\xi}} \bigl(\sqrt{2M}-1 \bigr)(M+2)
\]
for $M,\xi> 0$. Then for any $\varepsilon\geq C(M,\xi)\sqrt{\frac
{\log
d + \log^2 n}{n^{1/2}}}$ and sufficiently large $n$, we have
\begin{eqnarray*}
&&\mathbb{P} \Bigl(\max_{jk} | S_n(\tilde{f})_{jk} - S_n(f)_{jk} |>
\varepsilon\Bigr) \\[-2pt]
&&\quad\leq\frac{c_{1}d}{(n\varepsilon^{2})^{2\xi}} +
\frac{c_{2}d}{n^{M\xi-1}} + c_{3}\exp
\biggl(-\frac{c_{4}n^{1/2}\varepsilon^2}{\log d + \log^2 n} \biggr),
\end{eqnarray*}
where $c_{1}, c_{2}, c_{3}, c_{4}$ are positive constants.\vadjust{\goodbreak}
\end{theorem}

The proof of this result involves a detailed Gaussian tail analysis,
and is given in \citet{npn:09}.

Using Theorem \ref{thm.keylemma} and the results of \citet{Rothman:08},
it can then be shown that the precision matrix is estimated at the
following rates in the Frobenius norm and the $\ell_{2}$-operator norm:
\[
\label{eq.Frobenius}
\|\hat{\Omega}_n -\Omega_0\|_{\mathrm{F}} = O_{P} \biggl( \sqrt
{\frac{(s +
d)\log d + \log^2 n}{n^{1/2}} } \biggr)
\]
and
\[
\label{eq.L2}
\|\hat{\Omega}_n -\Omega_{0}\|_{2} = O_{P} \biggl( \sqrt{\frac
{s\log d +
\log^2 n}{n^{1/2}} } \biggr),
\]
where
\begin{eqnarray*}
s \equiv\operatorname{Card} \bigl( \{(i,j) &\in&\{1,\ldots, d \}
\{1,\ldots, d \} |\\
&&\hspace*{8pt}{}\Omega_0(i,j)\neq0, i\neq j \} \bigr)
\end{eqnarray*}
is the number of nonzero off-diagonal elements of the true precision
matrix.

Using the results of \citet{Ravikumar:Gauss:09}, it can also be shown,
under appropriate conditions, that the sparsity pattern of the
precision matrix is estimated accurately with high probability. In
particular, the nonparanormal estimator $\hat{\Omega}_n$ satisfies
\[
\mathbb{P} ( \cG(\hat{\Omega}_n, \Omega_{0} ) ) \geq1-o(1),
\]
where $\cG( \hat{\Omega}_n, \Omega_{0} )$ is the event
\[
\bigl\{ \operatorname{sign} (\hat{\Omega}_n(j,k) ) = \operatorname{sign}
(\Omega_{0}(j,k) ) ,\forall j,k \in\{1,\ldots, d\}
\bigr\}.
\]
We refer to \citet{npn:09} for the details of the conditions and proofs.
These $\tilde O_P(n^{-1/4})$ rates are slower than the $\tilde
O_P(n^{-1/2})$ rates obtainable for the graphical lasso. However, in
more recent work (\citep{Liu:12}) we use estimators based on Spearman's
rho and Kendall's tau statistics to obtain the parametric rate.

\section{Forest Density Estimation}
\label{sec.notation}

We now describe a very different, but equally flexible and useful
approach. Rather than assuming a transformation to normality and an
arbitrary undirected graph, we restrict the graph to be a tree or
forest, but allow arbitrary nonparametric distributions.

Let $\truep(x)$ be a probability density with respect to Lebesgue
measure $\mu(\cdot)$ on $\mathbb{R}^{d}$, and let $X^{(1)}, \ldots,
X^{(n)}$ be $n$ independent identically distributed
$\mathbb{R}^{d}$-valued data vectors sampled from $\truep(x)$ where
$X^{(i)} = (X^{(i)}_1,\break\ldots, X^{(i)}_d)$. Let $\X_{j}$ denote the
range of $X_j^{(i)}$, and let $\X= \X_{1} \times\cdots\times
\X_{d}$.

A graph is a forest if it is acyclic. If $F$ is a $d$-node undirected
forest with vertex set $V_{F} = \{1,\ldots,d \}$ and edge set $E_{F}
\subset\{1, \ldots, d \} \times\{1,\ldots, d\}$, the number of edges
satisfies $|E_{F}|<d$. We say that a probability density function
$p(x)$ is \textit{supported by a forest $F$} if the density can be
written as
%
\begin{equation}\label{eq.treedensity}
p_F(x) = \prod_{(i,j)\in E_{F}} \frac{ p(x_{i},
x_{j})}{p(x_{i})p(x_{j})} \prod_{k\in V_{F}}p(x_{k}),
\end{equation}
where each $p(x_{i}, x_{j})$ is a bivariate density on $\X_i\times
X_j$, and each $p(x_{k})$ is a univariate density on $\X_k$.

Let $\mathcal{F}_{d}$ be the family of forests with $d$ nodes, and let
$\mathcal{P}_{d}$ be the corresponding family of densities.
\begin{eqnarray}\label{eq.Pd}
\qquad\mathcal{P}_{d} &=& \biggl\{ p\geq0\dvtx\int_{\X} p(x)\,
d\mu(x) = 1,\mbox{ and}\nonumber
\\[-8pt]\\[-8pt]
&&\hphantom{\{}{} \mbox{$p(x)$
satisfies \eqref{eq.treedensity} for some $F\in\mathcal{F}_d$}
\biggr\}.\nonumber
\end{eqnarray}
Define the oracle forest density
%
\begin{equation}\label{eq.oracle}
q^{*}= \argmin\limits_{q \in\mathcal{P}_{d}} D(\truep\div q)
\end{equation}
where the Kullback--Leibler divergence $D(p\div q)$ between two
densities $p$ and $q$ is
%
\begin{equation}
D(p\div q) = \int_{\X} p(x) \log\frac{p(x)}{q(x)} \,dx,
\end{equation}
under the convention that $0\log(0/q) = 0$, and\break $p\log(p/0) =
\infty$
for $p\neq0$. The following is straightforward to prove.

\begin{proposition}\label{prop.oracle}
Let $q^{*}$ be defined as in \eqref{eq.oracle}. There exists a forest
$F^{*}\in\mathcal{F}_{d}$, such that
\begin{eqnarray}\label{eq.Tstar}
q^{*} &=& \truep_{F^{*}}\nonumber
\\[-8pt]\\[-8pt] &=& \prod_{(i,j)\in E_{F^{*}}}\frac{ \truep(x_{i},
x_{j})}{\truep(x_{i})\truep(x_{j})} \prod_{k\in
V_{F^{*}}}\truep(x_{k}),\nonumber
\end{eqnarray}
where $\truep(x_{i}, x_{j})$ and $\truep(x_{i})$ are the bivariate and
univariate marginal densities of $\truep$.
\end{proposition}

For any density $q(x)$, the negative log-likelihood risk $R(q)$ is
defined as
\begin{eqnarray}
R(q) &=& -\mathbb{E}\log q(X)\nonumber
\\[-8pt]\\[-8pt] &=& -\int_{\X} \truep(x)\log q(x)\, dx.\nonumber
\end{eqnarray}
It is straightforward to see that the density $q^{*}$ defined in
\eqref{eq.oracle} also minimizes\vadjust{\goodbreak} the negative log-likeli\-hood loss.
\begin{eqnarray}
q^{*} &=& \argmin\limits_{q\in\mathcal{P}_d} D(\truep\div
q)\nonumber
\\[-8pt]\\[-8pt] &=& \argmin\limits_{q
\in
\mathcal{P}_{d}}R(q).\nonumber
\end{eqnarray}

We thus define the oracle risk as $R^{*}=R(q^{*})$. Using
Proposition \ref{prop.oracle} and equation \eqref{eq.treedensity}, we
have
\begin{eqnarray}\label{eq.oraclerisk}
\nonumber
R^{*} &=& R(q^{*}) = R(\truep_{F^{*}})\nonumber\\
& = & -\int_{\X} \truep(x) \biggl( \sum_{(i,j)\in E_{F^{*}}} \log
\frac{
\truep(x_{i}, x_{j})}{\truep(x_{i})\truep(x_{j})}\nonumber
\\[-8pt]\\[-8pt]
&&\hspace*{83pt}{} +
\sum_{k\in V_{F^{*}}}\log( \truep(x_{k}) ) \biggr)\,dx \nonumber\\
& = &-\sum_{(i,j)\in E_{F^{*}}} I(X_{i}; X_{j}) + \sum_{k\in
V_{F^{*}}} H(X_{k}),\nonumber
\end{eqnarray}
where
\begin{eqnarray}
\quad I(X_{i}; X_{j})
&= &\int_{\X_{i}\times\X_{j}} \truep(x_{i}, x_{j})\nonumber
\\[-8pt]\\[-8pt]
&& \hphantom{\int_{\X_{i}\times\X_{j}}}{}\cdot\log
\frac{\truep(x_{i}, x_{j}) }{\truep(x_{i}) \truep(x_{j})} \,dx_{i}\,
dx_{j}\nonumber
\end{eqnarray}
is the mutual information between the pair of variables $X_{i}$,
$X_{j}$, and
%
\begin{equation}
H(X_{k}) = -\int_{\X_{k}} \truep(x_{k}) \log\truep(x_{k}) \,dx_{k}
\end{equation}
is the entropy.

\subsection{A Two-Step Procedure}
\label{method}

If the true density $\truep(x)$ were known, by Proposition~\ref{prop.oracle}, the density estimation problem would be reduced to
finding the best forest structure $F^{*}_{d}$, satisfying
\begin{eqnarray}
F^{*}_{d} &=& \argmin\limits_{F\in\mathcal{F}_{d}}R(\truep
_{F})\nonumber
\\[-8pt]\\[-8pt]& =&
\argmin\limits_{F\in\mathcal{F}_{d}} D(\truep\div\truep
_{F}).\nonumber
\end{eqnarray}
The optimal forest $F^{*}_{d}$ can be found by minimizing the
right-hand side of \eqref{eq.oraclerisk}. Since the entropy term $H(X)
= \sum_{k} H(X_{k})$ is constant across all forests, this can be recast
as the problem of finding the maximum weight spanning forest for a
weighted graph, where the weight of the edge connecting nodes $i$ and
$j$ is $I(X_{i}; X_{j})$. Kruskal's algorithm
(\citep{Kruskal:1956}) is a greedy algorithm that is guaranteed to find
a maximum weight spanning\vadjust{\goodbreak} tree of a weighted graph. In the setting of
density estimation, this procedure was proposed by \citet{chow68} as a
way of constructing a tree approximation to a distribution. At each
stage the algorithm adds an edge connecting that pair of variables with
maximum mutual information among all pairs not yet visited by the
algorithm, if doing so does not form a cycle. When stopped early, after
$k<d-1$ edges have been added, it yields the best $k$-edge weighted
forest.

Of course, the above procedure is not practical since the true density
$\truep(x)$ is unknown. We replace the population mutual information
$I(X_{i}; X_{j})$ in \eqref{eq.oraclerisk} by a plug-in estimate
$\hat{I}_n(X_{i}; X_{j})$, defined as
\begin{eqnarray}
\hat{I}_n(X_{i}; X_{j}) &=& \int_{\X_{i}\times\X_{j}} \hat{p}_n(x_{i},
x_{j}) \nonumber
\\[-8pt]\\[-8pt]
&&\hphantom{\int_{\X_{i}\times\X_{j}}}{}\cdot\log\frac{\hat
{p}_n(x_{i}, x_{j}) }{\hat{p}_n(x_{i})
\hat{p}_n(x_{j})} \,dx_{i}\,dx_{j},\nonumber
\end{eqnarray}
where $\hat{p}_n(x_{i}, x_{j})$ and $\hat{p}_n(x_{i})$ are bivariate
and univariate kernel density estimates. Given this estimated mutual
information matrix $\hat{M}_n = [\hat{I}_n(X_{i}; X_{j}) ]$, we can
then apply Kruskal's algorithm (equivalently, the Chow--Liu
algorithm) to find the best tree structure $\hat{F}_{n}$.

Since the number of edges of $\hat{F}_{n}$ controls the number of
degrees of freedom in the final density estimator, an automatic
data-dependent way to choose it is needed. We adopt the following
two-stage procedure. First, we randomly split the data into two sets
$\mathcal{D}_{1}$ and $\mathcal{D}_{2}$ of sizes $n_{1}$ and $n_{2}$;
we then apply the following steps:

\begin{longlist}
\item[(1)] Using $\mathcal{D}_{1}$, construct kernel density estimates
of
the univariate and bivariate marginals and calculate
$\hat{I}_{n_1}(X_{i}; X_{j})$ for $i,j \in\{1, \ldots, d \}$ with
$i\neq j$. Construct a full tree $\hat{F}^{(d-1)}_{n_1}$ with $d-1$
edges, using the Chow--Liu algorithm.

\item[(2)] Using $\mathcal{D}_{2}$, prune the tree
$\hat{F}^{(d-1)}_{n_1}$
to find a forest $\hat{F}^{(\hat k)}_{n_1}$ with
$\hat{k}$ edges, for $0\leq\hat{k} \leq d-1$.
\end{longlist}

Once $\hat{F}^{(\hat k)}_{n_1}$ is obtained in Step 2, we can calculate
$\hat{p}_{\hat{F}^{(\hat k)}_{n_1}}$ according to
\eqref{eq.treedensity}, using the kernel density estimates constructed
in Step 1.

\subsubsection{Step 1: Constructing a sequence of forests} 

Step 1 is carried out on the dataset $\mathcal{D}_{1}$. Let $K(\cdot)$
be a univariate kernel function. Given an evaluation point $(x_{i},
x_{j})$, the bivariate kernel density estimate for $(X_{i}, X_{j})$
based on the observations $\{X^{(s)}_{i},\break
X^{(s)}_{j}\}_{s\in\mathcal{D}_{1}}$ is defined as
\begin{eqnarray}\label{eq.bivariatekde}
&&\qquad\hat{p}_{n_1}(x_{i}, x_{j})\nonumber
\\[-3pt]\\[-13pt]
&&\quad\qquad= \frac{1}{n_{1}}\sum_{s \in
\mathcal{D}_{1}} \frac{1}{h^{2}_{2}} K \biggl( \frac{X^{(s)}_{i} -
x_{i}}{h_{2}} \biggr) K \biggl( \frac{X^{(s)}_{j} - x_{j}}{h_{2}}
\biggr),\nonumber
\end{eqnarray}
where we use a product kernel with $h_{2} > 0$ as the bandwidth
parameter. The univariate kernel density estimate
$\hat{p}_{n_1}(x_{k})$ for $X_{k}$ is
%
\begin{equation}\label{eq.univariatekde}
\hat{p}_{n_1}(x_{k}) = \frac{1}{n_{1}}\sum_{s \in\mathcal{D}_{1}}
\frac{1}{h_{1}} K \biggl( \frac{X^{(s)}_{k} - x_{k}}{h_{1}} \biggr),
\end{equation}
where $h_{1}>0$ is the univariate bandwidth.

We assume that the data lie in a $d$-dimensional unit cube $\X=
[0,1]^{d}$. To calculate the empirical mutual information
$\hat{I}_{n_1}(X_{i}; X_{j})$, we need to numerically evaluate a
two-dimensional integral. To do so, we calculate the kernel density
estimates on a grid of points. We choose $m$ evaluation points on each
dimension, $x_{1i} < x_{2i} < \cdots< x_{mi}$ for the $i$th variable.
The mutual information $\hat{I}_{n_1}(X_{i}; X_{j})$ is then
approximated as
%
\begin{eqnarray}\label{eq.empiricalMI}
&&\hat{I}_{n_1}(X_{i}; X_{j})\nonumber
\\
&&\quad=
\frac{1}{m^{2}}\sum_{k=1}^{m}\sum_{\ell=1}^{m}\hat
{p}_{n_1}(x_{ki},x_{\ell
j})
\\
&&\qquad\hphantom{\frac{1}{m^{2}}\sum_{k=1}^{m}\sum_{\ell
=1}^{m}}{}\cdot\log\frac{\hat{p}_{n_1}(x_{ki}, x_{\ell j})
}{\hat{p}_{n_1}(x_{ki})
\hat{p}_{n_1}(x_{\ell j})}.\nonumber
\end{eqnarray}
The approximation error can be made arbitrarily small by choosing $m$
sufficiently large. As a practical concern, care needs to be taken
that the factors $\hat{p}_{n_1}(x_{ki})$ and $\hat{p}_{n_1}(x_{\ell
j})$ in the denominator are not too small; a truncation procedure can
be used to ensure this. Once the $d\times d$ mutual information matrix
$\hat M_{n_1} = [\hat{I}_{n_1}(X_{i}; X_{j}) ]$ is obtained, we can
apply the Chow--Liu (Kruskal) algorithm to find a maximum weight
spanning tree (see Algorithm \ref{algorithm:naiveMI}).

\begin{algorithm}
\caption{Tree construction (Kruskal/Chow--Liu)}\label{algorithm:naiveMI}
Input: Data set $\mathcal{D}_{1}$ and the bandwidths $h_{1}$,
$h_{2}$.

\hangindent=12pt Initialize: Calculate $\hat M_{n_1}$, according to
\eqref{eq.bivariatekde},
\eqref{eq.univariatekde}  and \eqref{eq.empiricalMI}.

Set $E^{(0)}=\varnothing$.

For $k = 1, \ldots, d-1$:

\hangindent=12pt\mbox{\quad}(1) Set $(i^{(k)}, j^{(k)}) \leftarrow
\argmax_{(i,j)} \hat
M_{n_1}(i,j)$ such
that $E^{(k-1)} \cup\{(i^{(k)}, j^{(k)}) \}$ does not contain a cycle;

\hangindent=12pt\mbox{\quad}(2) $E^{(k)} \leftarrow E^{(k-1)} \cup\{
(i^{(k)}, j^{(k)})\}$.

Output: tree $\hat F^{(d-1)}_{n_1}$ with edge set $E^{(d-1)}$.
\end{algorithm}

\subsubsection{Step 2: Selecting a forest size}\label{subsubsec.stage2}

The full tree $\hat{F}^{(d-1)}_{n_1}$ obtained in Step 1 might have
high variance when the dimension $d$ is large, leading to overfitting
in the density estimate. In order to reduce the variance, we prune the
tree; that is, we choose an unconnected tree with $k$ edges. The
number of edges $k$ is a tuning parameter that induces a
bias--variance tradeoff.

In order to choose $k$, note that in stage $k$ of the Chow--Liu
algorithm, we have an edge set $E^{(k)}$ (in the notation of the
Algorithm~\ref{algorithm:naiveMI}) which corresponds to a forest $\hat
F^{(k)}_{n_1}$ with $k$ edges, where $F^{(0)}_{n_1}$ is the union of
$d$ disconnected nodes. To select $k$, we cross-validate over the $d$
forests $\hat F^{(0)}_{n_1}, \hat F^{(1)}_{n_1}, \ldots, \hat
F^{(d-1)}_{n_1}$.

Let $\hat{p}_{n_2}(x_{i}, x_{j})$ and $\hat{p}_{n_2}(x_{k})$ be defined
as in \eqref{eq.bivariatekde} and \eqref{eq.univariatekde}, but now
evaluated solely based on the held-out data in $\mathcal{D}_{2}$. For a
density $p_{F}$ that is supported by a forest $F$, we define the
held-out negative log-likelihood risk as
\begin{eqnarray}\label{eq.heldoutrisk}
&&\hat{R}_{n_2}(p_{F})\nonumber\\
&&\quad= -\sum_{(i,j)\in E_{F}} \int_{\X_{i}\times\X_{j}}
\hat{p}_{n_2}(x_{i}, x_{j})\nonumber
\\[-8pt]\\[-8pt]&&\qquad\hphantom{-\sum_{(i,j)\in E_{F}} \int_{\X
_{i}\times\X_{j}}}{}\cdot\log\frac{ p(x_{i},
x_{j})}{p(x_{i})p(x_{j})} \,dx_{i}\,dx_{j}\nonumber
\\
&&\qquad-
\sum_{k\in V_{F}}\int_{\X_{k}}\hat{p}_{n_2}(x_{k})\log p(x_{k})
\,dx_{k}.\nonumber
\end{eqnarray}
The selected forest is then $\hat{F}^{(\hat k)}_{n_1}$ where
%
\begin{eqnarray}
\hat{k}=\argmin\limits_{k \in\{0, \ldots, d-1 \}}\hat{R}_{n_2}
(\hat{p}_{F^{(k)}_{n_1}} )
\end{eqnarray}
and where $\hat{p}_{F^{(k)}_{n_1}}$ is computed using the density
estimate $\hat{p}_{n_1}$ constructed on $\mathcal{D}_1$.

We can also estimate $\hat{k}$ as
%
\begin{eqnarray}
\quad\hat{k} &=& \argmax\limits_{k \in\{0, \ldots, d-1 \}}
\frac{1}{n_{2}}\nonumber
\\
&&{}\cdot\sum_{s\in\mathcal{D}_{2}}\log\biggl( \prod_{(i,j)\in
E_{F^{(k)}}} \frac{ \hat{p}_{n_1}(X^{(s)}_{i},
X^{(s)}_{j})}{\hat{p}_{n_1}(X^{(s)}_{i})\hat{p}_{n_1}(X^{(s)}_{j})}
\\
&&\hspace*{103pt}{}\cdot
\prod_{\ell\in V_{F^{(k)}}}\hat{p}_{n_1}\bigl(X^{(s)}_{\ell}\bigr) \biggr)
\nonumber\\
&=& \argmax\limits_{k \in\{0, \ldots, d-1 \}}
\frac{1}{n_{2}}\nonumber
\\[-8pt]\\[-8pt]
&&{}\cdot\sum_{s\in\mathcal{D}_{2}}\log\biggl( \prod_{(i,j)\in
E_{F^{(k)}}} \frac{ \hat{p}_{n_1}(X^{(s)}_{i},
X^{(s)}_{j})}{\hat{p}_{n_1}(X^{(s)}_{i}) \hat{p}_{n_1}(X^{(s)}_{j})}
\biggr)\nonumber
.
\end{eqnarray}
This minimization can be efficiently carried out by iterating over the
$d-1$ edges in $\hat{F}^{(d-1)}_{n_1}$.

Once $\hat{k}$ is obtained, the final forest-based kernel density
estimate is given by
%
\begin{equation}\label{eq.tkde}
\qquad\hat{p}_{n}(x) = \prod_{(i,j)\in E^{(\hat k)}}\frac{
\hat{p}_{n_1}(x_{i}, x_{j})}{\hat{p}_{n_1}(x_{i})\hat{p}_{n_1}(x_{j})}
\prod_{k}\hat{p}_{n_1}(x_{k}).\hspace*{-20pt}
\end{equation}

Another alternative is to compute a maximum weight spanning
forest, using Kruskal's algorithm, but with held-out edge weights
%
\begin{equation}
\hspace*{20pt}\hat w_{n_2}(i,j) = \frac{1}{n_{2}}\sum_{s\in\mathcal
{D}_{2}} \log
\frac{ \hat{p}_{n_1}(X^{(s)}_{i},
X^{(s)}_{j})}{\hat{p}_{n_1}(X^{(s)}_{i})\hat
{p}_{n_1}(X^{(s)}_{j})}.\hspace*{-20pt}
\end{equation}
In fact, asymptotically (as $n_2\rightarrow\infty$) this gives an
optimal tree-based estimator constructed in terms of the kernel density
estimates $\hat p_{n_1}$.

\subsection{Statistical Properties}
\label{sec.theory}

The statistical properties of the forest density estimator can be
analyzed under the same type of assumptions that are made for classical
kernel density estimation. In particular, assume that the univariate
and bivariate densities lie in a H\"older class with exponent $\beta$.
Under this assumption the minimax rate of convergence in the
squared error loss is $O(n^{\beta/(\beta+1)})$ for bivariate densities
and $O(n^{2\beta/(2\beta+1)})$ for univariate densities. Technical
assumptions on the kernel yield $L_\infty$ concentration results on
kernel density estimation (\citep{Gine:2002}).

Choose the bandwidths $h_1$ and $h_2$ to be used in the one-dimensional
and two-dimensional kernel density estimates according to
%
\begin{eqnarray}
\label{eq:band1}
h_{1} &\asymp& \biggl(\frac{\log n}{n} \biggr)^{1/(1+2\beta)}, \\
\label{eq:band2} h_{2} &\asymp& \biggl(\frac{\log n}{n}
\biggr)^{1/(2+2\beta)}.
\end{eqnarray}
This choice of bandwidths ensures the optimal rate of convergence. Let
$\mathcal{P}^{(k)}_{d}$ be the family of $d$-dimensional densities that
are supported by forests with at most $k$ edges. Then
%
\begin{equation}\label{eq.nestedclass}
\mathcal{P}^{(0)}_{d} \subset\mathcal{P}^{(1)}_{d} \subset\cdots
\subset\mathcal{P}^{(d-1)}_{d}.
\end{equation}
Due to this nesting property,
\begin{eqnarray}
\inf_{q_{F} \in\mathcal{P}^{(0)}_{d}} R(q_{F}) &\geq&\inf_{q_{F}
\in
\mathcal{P}^{(1)}_{d}} R(q_{F})\nonumber
\\[-8pt]\\[-8pt] &\geq&\cdots\geq\inf_{q_{F} \in
\mathcal{P}^{(d-1)}_{d}} R(q_{F}).\nonumber
\end{eqnarray}
This means that a full spanning tree would generally be selected if we
had access to the true distribution. However, with access to finite
data to estimate the densities ($\hat p_{n_1}$), the optimal procedure
is to use fewer than $d-1$ edges. The following result analyzes the
excess risk resulting from selecting the forest based on the heldout
risk $\hat R_{n_2}$.

\begin{theorem}\label{thm.randompersistency}
Let $\hat{p}_{\hat{F}^{(k)}_{d}}$ be the estimate with
$|E_{\hat{F}^{(k)}_{d}}| = k$ obtained after the first $k$ iterations
of the Chow--Liu algorithm. Then under (omitted) technical assumptions
on the densities and kernel, for any $1\leq k\leq d-1$,
\begin{eqnarray}\label{eq.randomriskrate}
&&\quad R(\hat{p}_{\hat{F}^{(k)}_{d}}) -\inf_{q_{F} \in\mathcal
{P}^{(k)}_{d}} R(q_{F})\nonumber\\[-8pt]\\[-8pt]
&&\qquad= O_{P} \biggl(k \sqrt{\frac{\log n + \log d}{n^{\beta
/(1+\beta)}}} +
d\sqrt{\frac{\log n + \log d}{n^{2\beta/(1+2\beta)}}} \biggr
)\nonumber
\end{eqnarray}
and
%
\begin{eqnarray}\label{eq.randomriskrate2}
&&R(\hat{p}_{\hat{F}^{(\hat{k})}_{d}}) - \min_{0\leq k \leq
d-1}R(\hat{p}_{\hat{F}^{(k)}_{d}})\nonumber\\
&&\quad= O_{P} \Biggl((k^* + \hat{k})\sqrt{\frac{\log n + \log
d}{n^{\beta/(1+\beta)}}}
\\
&&\hspace*{43pt}\qquad{}+ d\sqrt{\frac{\log n + \log
d}{n^{2\beta/(1+2\beta)}}} \Biggr),\nonumber
\end{eqnarray}
where $\hat k=\argmin_{0\leq k \leq d-1}\hat
R_{n_2}(\hat{p}_{\hat{F}^{(k)}_{d}})$ and $k^*=\break\argmin_{0\leq
k \leq
d-1}R(\hat{p}_{\hat{F}^{(k)}_{d}})$.
\end{theorem}

The main work in proving this result lies in establishing bounds such
as
\begin{eqnarray}
&&\sup_{F\in\mathcal{F}^{(k)}_{d}}|R(\hat{p}_{F}) -
\hat{R}_{n_2}(\hat{p}_{F}) |\nonumber
\\[-8pt]\\[-8pt]
&&\quad= O_{P} \bigl(\phi_{n}(k) + \psi_{n}(d)
\bigr),\nonumber
\end{eqnarray}
where $\hat{R}_{n_2}$ is the held-out risk, under the notation
%
\begin{eqnarray}
\phi_{n}(k) &=& k\sqrt{\frac{\log n + \log d}{n^{\beta/(\beta
+1)}}},\\
\psi_{n}(d) &=& d\sqrt{\frac{\log n + \log d}{n^{2\beta/(1+2\beta)}}}.
\end{eqnarray}
For the proof of this and related results, see \citet{fde:11}. Using
this, one easily obtains
%
\begin{eqnarray}
&&R(\hat{p}_{\hat{F}^{(\hat{k})}_{d}}) -
R(\hat{p}_{\hat{F}^{(k^{*})}_{d}})\nonumber\\
&&\quad= R(\hat{p}_{\hat{F}^{(\hat{k})}_{d}}) - \hat{R}_{n_2}(\hat
{p}_{\hat{F}^{(\hat{k})}_{d}})
\\
&&\qquad{}+ \hat{R}_{n_2}(\hat{p}_{\hat
{F}^{(\hat{k})}_{d}}) - R(\hat{p}_{\hat{F}^{(k^{*})}_{d}}) \nonumber
\\
&&\quad= O_{P}\bigl(\phi_{n}(\hat{k}) + \psi_{n}(d)\bigr
)\nonumber\\[-8pt]\\[-8pt]
&&\qquad{}+ \hat{R}_{n_2}(\hat
{p}_{\hat{F}^{(\hat{k})}_{d}}) - R(\hat{p}_{\hat
{F}^{(k^{*})}_{d}})\nonumber
\\
\label{eq.randomkey1}&&\quad\leq O_{P}\bigl(\phi_{n}(\hat{k}) +
\psi_{n}(d)\bigr)\nonumber
\\[-8pt]\\[-8pt]
&&\qquad{}+ \hat
{R}_{n_2}(\hat{p}_{\hat{F}^{(k^{*})}_{d}}) - R(\hat{p}_{\hat
{F}^{(k^{*})}_{d}}) \nonumber\\
\label{eq.randomkey2}&&\quad= O_{P} \bigl(\phi_{n}(\hat{k}) + \phi
_{n}(k^{*}) + \psi_{n}(d) \bigr),
\end{eqnarray}
where \eqref{eq.randomkey1} follows from the fact that $\hat{k}$ is the
minimizer of $\hat{R}_{n_2}(\cdot)$. This result allows the dimension
$d$ to increase at a rate $ o ( \sqrt{ n ^{2\beta/(1+2\beta)} /\log n}
)$, and the number of edges $k$ to increase at a rate $ o (
\sqrt{ n ^{\beta/(1+\beta)} /\log n} )$, with the excess risk still
decreasing to zero asymptotically.

Note that the minimax rate for 2-dimensional kernel density estimation
under our stated conditions is $n^{-\beta/(\beta+1)}$. The rate above
is essentially the square root of this rate, up to logarithmic factors.
This is because a higher order kernel is used, which may result in
negative values. Once we correct these negative values, the resulting
estimated density will no longer integrate to one. The slower rate is
due to a very simple truncation technique to correct the higher-order
kernel density estimator to estimate mutual information. Current work
is investigating a different version of the higher order kernel density
estimator with more careful correction techniques, for which it is
possible to achieve the optimal minimax rate.

\begin{figure}

\includegraphics{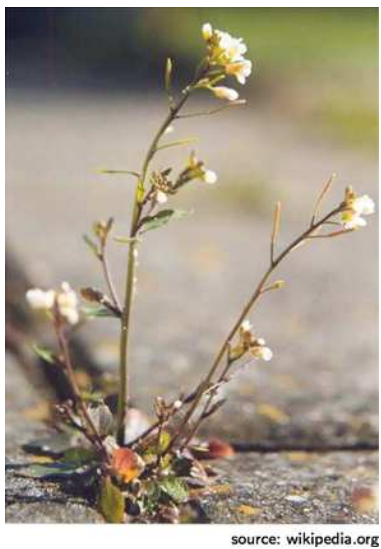}

\caption{\textup{Arabidopsis thaliana} is
a small
flowering plant; it was the first
plant genome to be sequenced, and its roughly 27,000 genes and
35,000 proteins have been actively studied. Here we consider a data
set based on Affymetrix GeneChip microarrays with sample size
$n=118$, for which $d=40$ genes have been selected for analysis.}\label{f4}
\end{figure}

\begin{figure*}[b]

\includegraphics{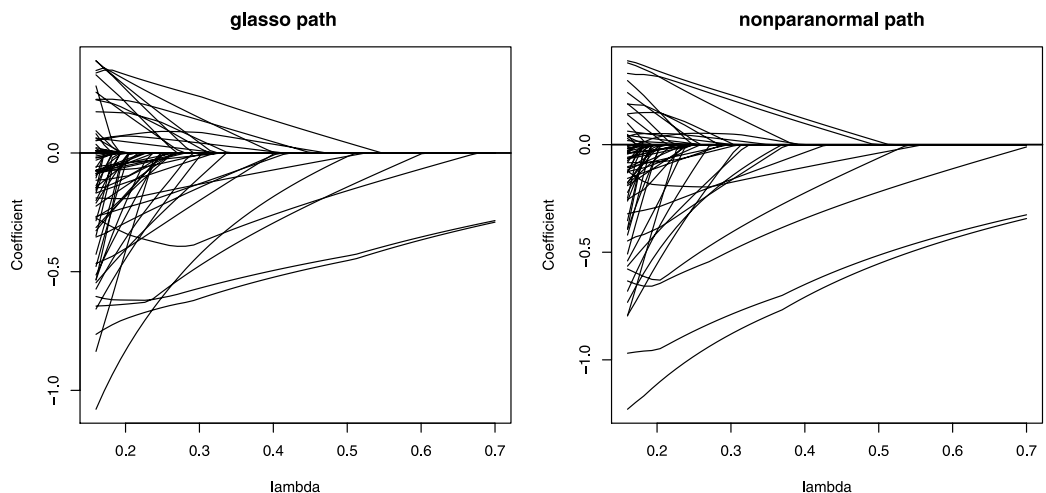}

\caption{Regularization paths of both methods on the microarray data
set. Although the paths for the two methods look similar, there are
some subtle differences.}\label{fig.GeneFullPath}
\end{figure*}

In theory the bandwidths are chosen as in \eqref{eq:band1} and
\eqref{eq:band2}, assuming $\beta$ is known. In our experiments
presented below, the bandwidth $h_{k}$ for the 2-dimensional kernel
density estimator is chosen according to the Normal reference rule
\begin{eqnarray}
h_{k} &=& 1.06\cdot\min\biggl\{\hat{\sigma}_{k},
\frac{\hat{q}_{k,0.75} - \hat{q}_{k,0.25}}{1.34} \biggr\}\nonumber
\\[-8pt]\\[-8pt]
&&{}\cdot
n^{-1/(2\beta+2)},\nonumber
\end{eqnarray}
where $\hat{\sigma}_{k}$ is the sample standard deviation of
$\{X^{(s)}_{k}\}_{s \in\mathcal{D}_{1}}$, and $\hat{q}_{k, 0.75}$,
$\hat{q}_{k, 0.25}$ are the $75\%$ and $25\%$ sample quantiles of
$\{X^{(s)}_{k}\}_{s \in\mathcal{D}_{1}}$, with $\beta=2$. See
\citet{wasserman:2006} for a discussion of this choice of bandwidth.

\section{Examples}
\label{sec:experiments}

\subsection{Gene--Gene Interaction Graphs}

The nonparanormal and Gaussian graphical model can construct very
different graphs. Here we consider a data set based on Affymetrix
GeneChip microarrays for the plant \textit{Arabidopsis thaliana}
(Wille et~al., \citeyear{wille:04}) (see Figure~\ref{f4}). The sample size is $n=118$.
The expression levels
for each chip are pre-processed by log-transformation and
standardization. A subset of 40 genes from the isoprenoid pathway is
chosen for analysis.

While these data are often treated as multivariate Gaussian, the
nonparanormal and the glasso give very different graphs over a wide
range of regularization parameters, suggesting that the nonparametric
method could lead to different biological conclusions.

The regularization paths of the two methods are compared in
Figure~\ref{fig.GeneFullPath}. To generate the paths, we select 50
regularization parameters on an evenly spaced grid in the interval
$[0.16, 1.2]$. Although the paths for the two methods look similar,
there are some subtle differences. In particular, variables become
nonzero in a different order.

\begin{figure*}

\includegraphics{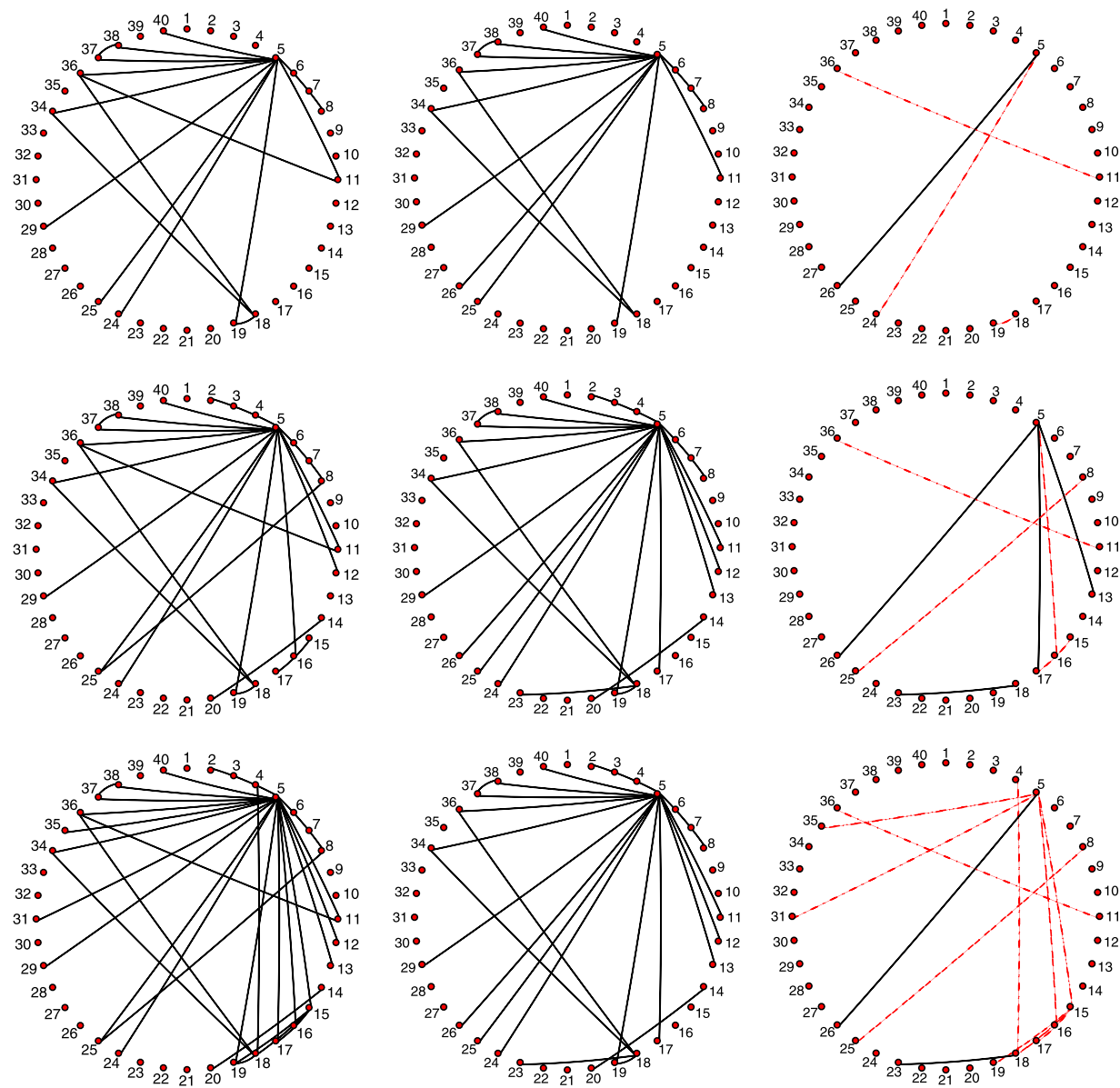}

\caption{The nonparanormal estimated graph for three values of
$\lambda=0.2448, 0.2661, 0.30857$ (left column), the closest glasso
estimated graph from the full path (middle) and the symmetric
difference graph (right). }
\label{fig.GeneSelectedGraph1}
\end{figure*}

\begin{figure*}

\includegraphics{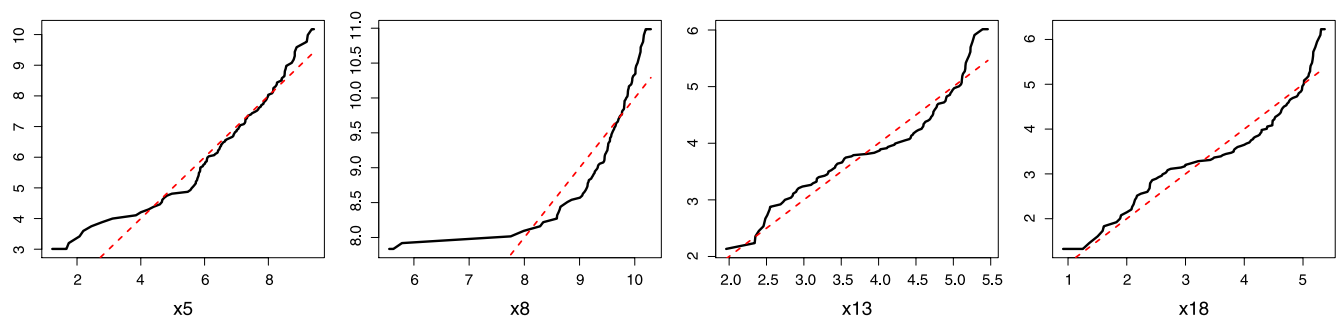}

\caption{Estimated transformation functions for four genes in the
microarray data set, indicating non-Gaussian marginals. The
corresponding genes are among the nodes appearing in the symmetric
difference graphs above.} \label{fig.Genecomponent1}
\end{figure*}

\begin{figure*}

\includegraphics{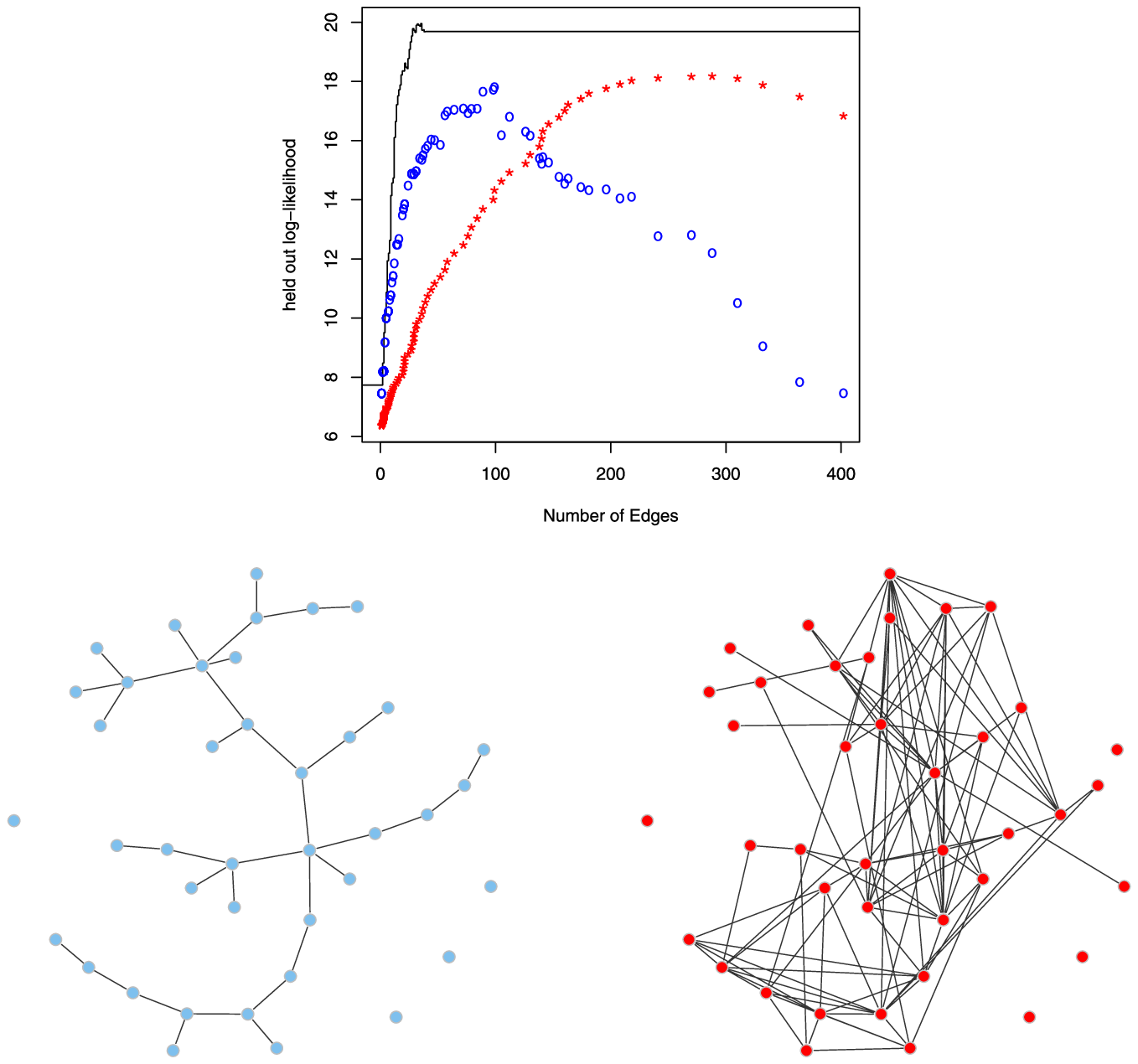}

\caption{Results on microarray data. Top: held-out
log-likelihood of the forest density estimator (black step function),
glasso (red stars)
and refit glasso (blue circles). Bottom: estimated graphs
using the forest-based estimator (left) and the glasso (right),
using the same node layout.}
\label{fig.loglikegene}
\end{figure*}

Figure \ref{fig.GeneSelectedGraph1} compares the estimated graphs for
the two methods at several values of the regularization parameter
$\lambda$ in the range $[0.16, 0.37]$. For each $\lambda$, we show the
estimated graph from the nonparanormal in the first column. In the
second column we show the graph obtained by scanning the full
regularization path of the glasso fit and finding the graph having
the smallest symmetric difference with the nonparanormal graph. The
symmetric difference graph is shown in the third column. The closest
glasso fit is different, with edges selected by the glasso not selected
by the nonparanormal, and vice-versa. The estimated transformation
functions for several genes are shown Figure \ref{fig.Genecomponent1},
which show non-Gaussian behavior.

Since the graphical lasso typically results in a large parameter bias
as a consequence of the $\ell_1$ regularization, it sometimes make
sense to use the \textit{refit glasso}, which is a two-step
procedure---in the first step, a~sparse inverse covariance matrix is
obtained by the graphical lasso; in the second step, a Gaussian model
is refit without $\ell_1$ regularization, but enforcing the sparsity
pattern obtained in the first step.

Figure~\ref{fig.loglikegene} compares forest density estimation to the
graphical lasso and refit glasso. It can be seen that the forest-based
kernel density estimator has better generalization performance. This
is not surprising, given that the true distribution of the data is not
Gaussian. (Note that since we do not directly compute the marginal
univariate densities in the nonparanormal, we are unable to compute
likelihoods under this model.) The held-out log-likelihood curve for
forest density estimation achieves a maximum when there are only 35
edges in the model. In contrast, the held-out log-likelihood curves of
the glasso and refit glasso achieve maxima when there are around 280
edges and 100 edges respectively, while their predictive estimates are
still inferior to those of the forest-based kernel density estimator.
Figure \ref{fig.loglikegene} also shows the estimated graphs for the
forest-based kernel density estimator and the graphical lasso. The
graphs are automatically selected based on held-out log-likelihood, and
are clearly different.

\begin{figure*}
\begin{tabular*}{\textwidth}{@{\extracolsep{\fill}}cc}
\begin{tabular}{ll}
Target Corp. & (\texttt{Consumer Discr.})\\
\hline
Big Lots, Inc. & (\texttt{Consumer Discr.})\\
Costco Co. & (\texttt{Consumer Staples})\\
Family Dollar Stores & (\texttt{Consumer Discr.})\\
Kohl's Corp. & (\texttt{Consumer Discr.})\\
Lowe's Cos. & (\texttt{Consumer Discr.})\\
Macy's Inc. & (\texttt{Consumer Discr.})\\
Wal-Mart Stores & (\texttt{Consumer Staples})
\end{tabular}
&
\begin{tabular}{ll}
Yahoo Inc. & (\texttt{Information Tech.}) \\
\hline
Amazon.com Inc. & (\texttt{Consumer Discr.})\\
eBay Inc. & (\texttt{Information Tech.})\\
NetApp & (\texttt{Information Tech.})
\end{tabular}
\end{tabular*}
\caption{Example neighborhoods in a forest graph for two stocks, Yahoo
Inc.
and Target Corp. The corresponding GICS industries are shown in
parentheses. (\texttt{Consumer Discr.}~is short for \texttt{Consumer
Discretionary}, and
\texttt{Information Tech.}~is short for \texttt{Information Technology}.)}\label{nbhds}
\end{figure*}

\subsection{Graphs for Equities Data}
\label{sec:stocks}

For the examples in this section we collected stock price data from
Yahoo! Finance (\href{http://finance.yahoo.com}{finance.yahoo.com}). The daily closing prices
were obtained for 452 stocks that consistently were in the S\&P 500
index between January 1, 2003 through January 1, 2011. This gave us
altogether 2015 data points, each data point corresponds to the vector
of closing prices on a trading day. With $S_{t,j}$ denoting the
closing price of stock $j$ on day $t$, we consider the variables
$X_{tj} = \log(S_{t,j}/ S_{t-1,j} )$ and build graphs over the indices
$j$. We simply treat the instances $X_t$ as independent replicates,
even though they form a time series. The data contain many outliers;
the reasons for these outliers include splits in a stock, which
increases the number of shares. We Winsorize (or truncate) every stock
so that its data points are within three times the mean absolute
deviation from the sample average. The importance of this
Winsorization is shown below; see the ``snake graph'' in
Figure~\ref{fig.stock1}. For the following results we use the subset
of the data between January 1, 2003 to January 1, 2008, before the
onset of the ``financial crisis.'' It is interesting to compare to
results that include data after 2008, but we omit these for brevity.


The 452 stocks are categorized into 10 Global Industry Classification
Standard (GICS) sectors, including \texttt{ Consumer Discretionary} (70
stocks),\break \texttt{Consumer Staples} (35 stocks), \texttt{Energy} (37
stocks), \texttt{Financials} (74 stocks), \texttt{Health Care}
(46 stocks), \texttt{Industrials} (59 stocks), \texttt{Information
Tech-\break nology}~(64 stocks), \texttt{Materials} (29 stocks),
\texttt{Telecom\-munications Services} (6 stocks), and \texttt{Utilities}
(32~stocks). In the graphs shown below, the nodes are colored
according to the GICS sector of the corresponding stock. It is expected
that stocks from the same GICS sectors should tend to be clustered
together, since stocks from the same GICS sector tend to interact more
with each other. This is indeed this case; for example, Figure~\ref{nbhds} shows
examples of the neighbors of two stocks, Yahoo Inc. and Target Corp.,
in the forest density graph.


\begin{figure*}

\includegraphics{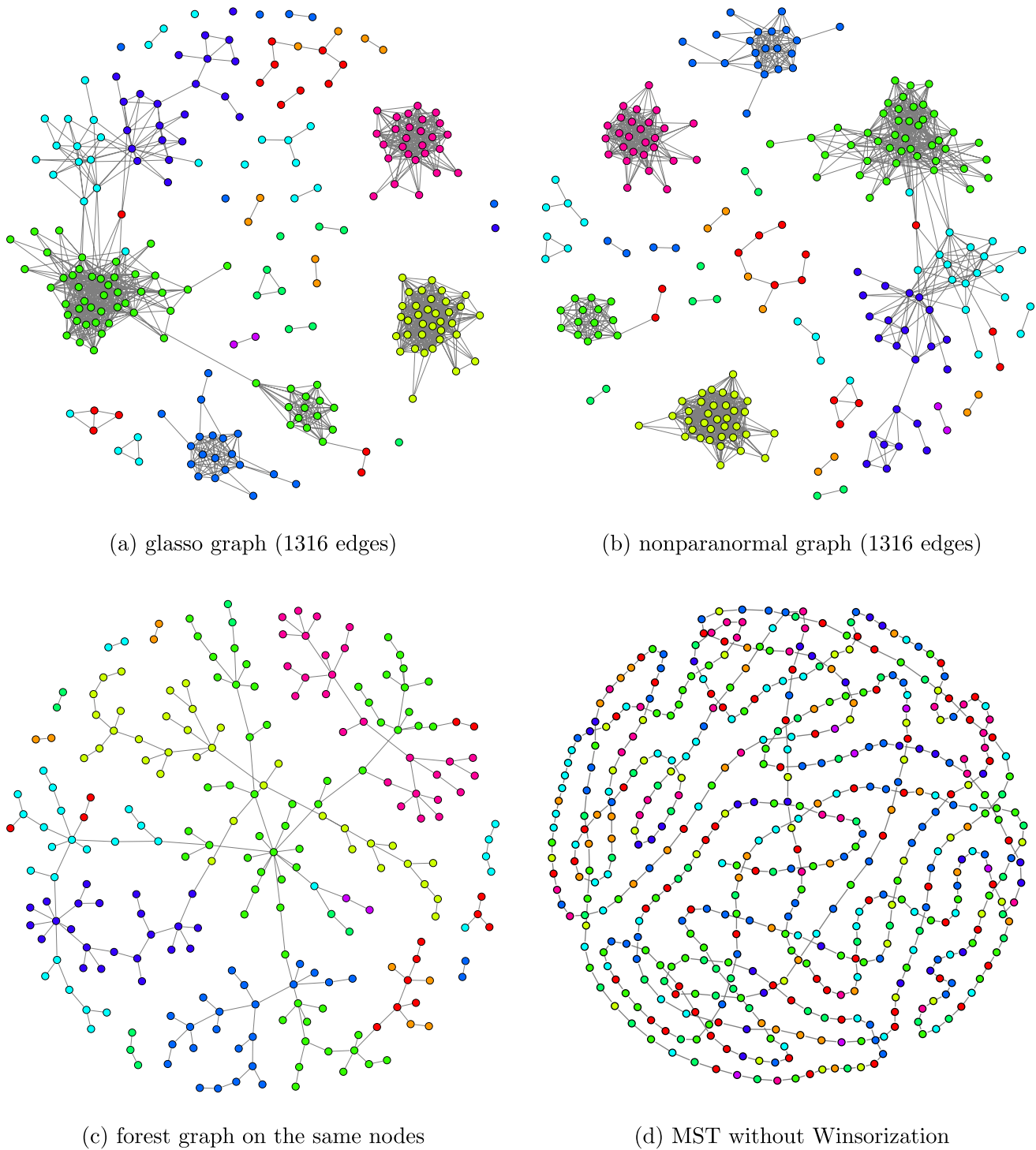}

\caption{Graphs build on S\&P 500 stock data from Jan.~1, 2003
to
Jan.~1, 2008. The graphs are estimated using \textup{(a)} the glasso,
\textup{(b)} the nonparanormal and \textup{(c)} forest density estimation. The nodes are
colored according to their GICS sector categories. Nodes are not
shown that have zero neighbors in both the glasso and nonparanormal
graphs. Figure \textup{(d)}
shows the maximum weight spanning tree that results if the data are not
Winsorized to trim outliers.}
\label{fig.stock1}
\end{figure*}

%

\begin{figure*}

\includegraphics{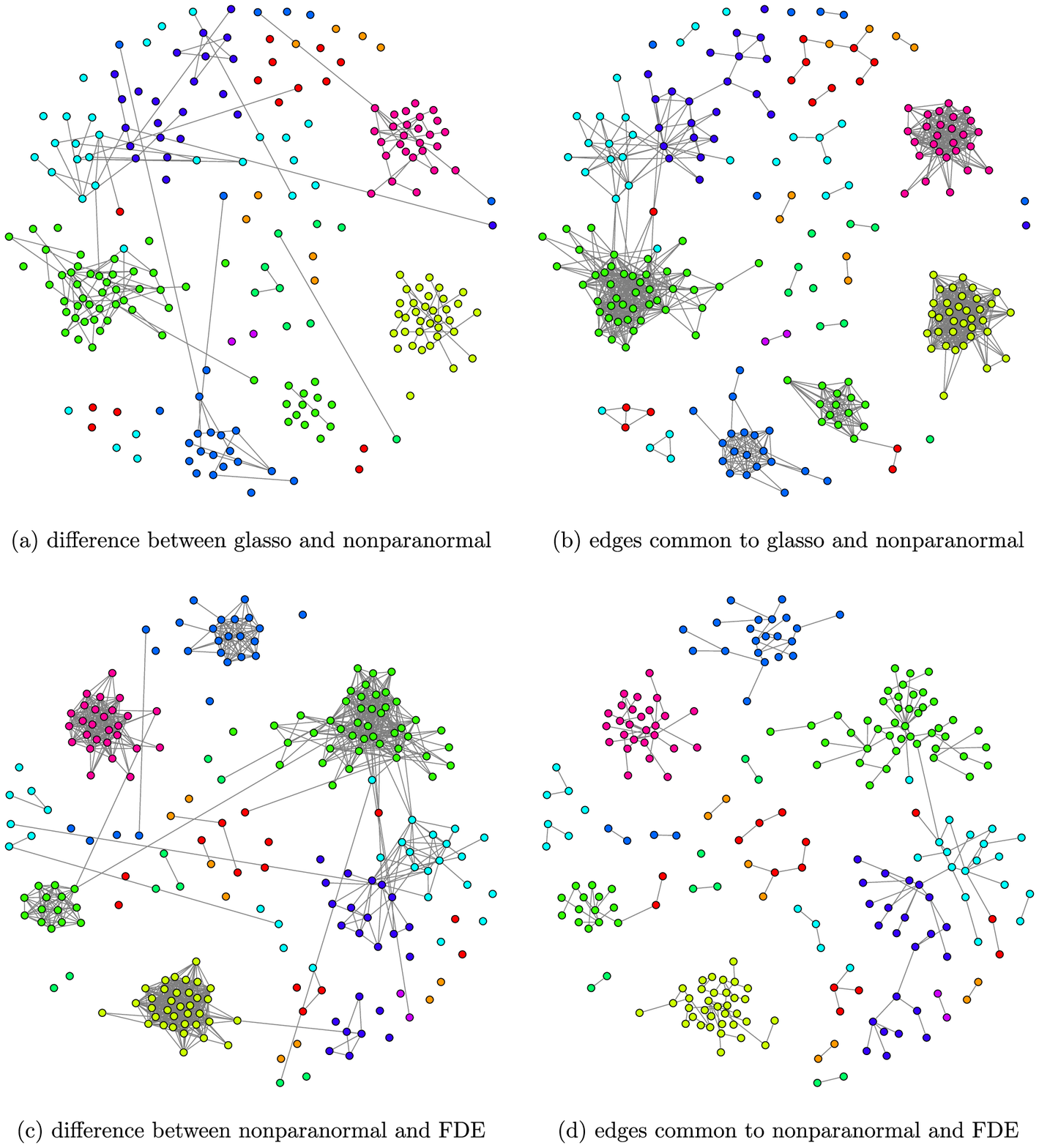}

\caption{Visualizations of the differences and similarities
between the
estimated graphs. The symmetric difference between the glasso and
nonparanormal graphs is shown in \textup{(a)}, and the edges common to the
graphs are shown in \textup{(b)}. Similarly, the symmetric difference between
the nonparanormal and forest density estimate is shown in \textup{(c)}, and the
common edges are shown in \textup{(d)}.}
\label{fig.stock1a}
\end{figure*}

Figures \ref{fig.stock1}(a)--(c) show graphs estimated using the
glasso, nonparanormal, and forest density estimator on the data from
January 1, 2003 to January 1, 2008. There are altogether $n=1257$
data points and $d=452$ dimensions. To estimate the glasso graph, we
somewhat arbitrarily set the regularization parameter to $\lambda=0.55$,
which results in a graph that has 1316 edges, about 3 neighbors per
node, and good clustering structure. The resulting graph is shown in
Figure~\ref{fig.stock1}(a). The corresponding nonparanormal graph is
shown in Figure~\ref{fig.stock1}(b). The regularization is chosen so
that it too has 1316 edges. Only nodes that have neighbors in one of
the graphs are shown; the remaining nodes are disconnected.

Since our dataset contains $n=1257$ data points, we directly apply
the forest density estimator on the whole dataset to obtain a full
spanning tree of $d-1=451$ edges. This estimator turns out to be very
sensitive to outliers, since it exploits kernel density estimates as
building blocks. In Figure \ref{fig.stock1}(d) we show the estimated
forest density graph on the stock data when outliers are \textit{not}
trimmed by Winsorization. In this case the graph is anomolous, with a
snake-like character that weaves in and out of the 10 GICS industries.
Intuitively, the outliers make the two-dimensional densities appear
like thin ``pancakes,'' and densities with similar orientations are
clustered together. To address this, we trim the outliers by
Winsorizing at 3 MADs, as described above. Figure \ref{fig.stock1}(c)
shows the estimated forest graph, restricted to the same stocks shown
for the graphs in (a) and (b). The resulting graph has good clustering
with respect to the GICS sectors.

Figures \ref{fig.stock1a}(a)--(c) display the differences and edges
common to the glasso, nonparanormal and forest graphs. Figure
\ref{fig.stock1a}(a) shows the symmetric difference between the
estimated glasso and nonparanormal graphs, and Figure
\ref{fig.stock1a}(b) shows the common edges. Figure
\ref{fig.stock1a}(c) shows the symmetric difference between the
nonparanormal and forest graphs, and Figure \ref{fig.stock1a}(d) shows
the common edges.

We refrain from drawing any hard conclusions\break about the effectiveness of
the different methods based on these plots---how these graphs are used
will depend on the application. These results serve mainly to
highlight how very different inferences about the independence
relations can arise from moving from a Gaussian model to a
semiparametric model to a fully nonparametric model with restricted
graphs.

\section{Related Work}

There is surprisingly little work on structure learning of
nonparametric graphical models in high dimensions. One piece of related
work is sparse log-density smoothing spline ANOVA models, introdu\-ced by
\citet{Jeon:Lin:2006}. In such a model the log-density function is
decomposed as the sum of a constant term, one-dimensional functions
(main effects), two-dimensional functions (two-way interactions) and so
on.
%
\begin{eqnarray}
\hspace*{18pt}\log p(x) &=& f(x)\hspace*{-18pt} \nonumber\\
[-3pt]\\[-15pt]
&\equiv& c + \sum_{j=1}^{d}f_{j}(x_{j}) +
\sum_{j<k}f_{jk}(x_{j}, x_{k}) + \cdots.\nonumber
\end{eqnarray}
The component functions satisfy certain constraints so that the model
is identifiable. In high dimensions, the model is truncated up to
second order interactions so that the computation is still tractable.
There is a close connection between the log-density ANOVA model and
undirected graphical models. For a model with only main effects and
two-way interactions, we define a graph $G=(V, E)$ such that $(i,j)\in
E$ if and only if $f_{ij}\neq0$. It can be seen that $p(x)$ is Markov
to~$G$. \citet{Jeon:Lin:2006} assume that these component functions
belong to certain reproducing kernel Hilbert spaces (RKHSs) equipped
with a RKHS norm $\|\cdot\|_{K}$. To obtain a sparse estimation of the
component functions $f(x)$, they propose a penalized M-estimator,
\begin{eqnarray} \label{eq::logdensity}
\hat{f} &=& \argmax\limits_{f} \Biggl\{ \frac{1}{n}\sum
_{i=1}^{n}\exp\bigl(f\bigl(X^{(i)}\bigr)
\bigr) \nonumber
\\[-8pt]\\[-8pt]
&&\hspace*{40pt}{}+ \int f(x)\rho(x)\,dx + \lambda J(f) \Biggr\},\nonumber
\end{eqnarray}
where $\rho(x)$ is some pre-defined positive density, and $J(f)$ is a
sparsity-inducing penalty that takes the form
%
\begin{equation}
J(f) = \sum_{j=1}^{d}\|f_{j} \|_{K} + \sum_{j<k}\|f_{jk} \|_{K}.
\end{equation}
Solving \eqref{eq::logdensity} only requires one-dimensional integrals
which can be efficiently computed. However, the optimization in
\eqref{eq::logdensity} exploits a surrogate loss instead of the
log-likelihood loss, and is more difficult to analyze theoretically.

Another related idea is to conduct structure learning using
nonparametric decomposable graphical\break \mbox{models}
(\citep{Schwaighofer:2007}). A distribution is a decomposable graphical
model if it is Markov to a graph $G=(V, E)$ which has a junction tree
representation, which can be viewed as an extension of tree-based
graphical models. A junction tree yields a factorized form
%
\begin{equation}
p(x) = \frac{\prod_{C\in V_{T}}p(x_{C})}{\prod_{S\in E_{T}} p(x_{S})},
\end{equation}
where $V_{T}$ denotes the set of cliques in $V$, and $E_{T}$ is the set
of separators, that is, the intersection of two neighboring cliques in
the junction tree. Exact search for the junction tree structure that
maximizes the likelihood is usually computationally expensive.
\citet{Schwaighofer:2007} propose a forward--back\-ward strategy for
nonparametric structure learning. However, such a greedy procedure
does not guarantee that the global optimal solution is found, and makes
theoretical analysis challenging.

\section{Discussion}

This paper has considered undirected graphical models for continuous
data, where the general densities take the form
%
\begin{equation}\label{eq:npgm2}
p(x) \propto\exp\biggl( \sum_{C\in\mathrm{Cliques}(G)}
f_C(x_C) \biggr).
\end{equation}
Such a general family is at least as difficult as the general
high-dimensional nonparametric regression model. But, as for
regression, simplifying assumptions can lead to tractable and useful
models. We have considered two approaches that make very different
tradeoffs between statistical generality and computational efficiency.
The nonparanormal relies on estimating one-dimensional functions, in a
manner that is similar to the way additive models estimate
one-dimensional regression functions. This allows arbitrary graphs,
but the distribution is semiparametric, via the Gaussian copula. At
the other extreme, when we restrict to acyclic graphs we can have fully
nonparametric bivariate and univariate marginals. This leverages
classical techniques for low-dimensional density estimation, together
with approximation algorithms for constructing the graph. Clearly
these are just two among many possibilities for nonparametric graphical
modeling. We conclude, then, with a brief description of a few
potential directions for future work.

As we saw with the nonparanormal, if only the graph is of interest, it
may not be important to estimate the functions accurately. More
generally, to estimate the graph it is not necessary to estimate the
density. One of the most effective and theoretically well-supported
methods for estimating Gaussian graphs is due to
\citet{Meinshausen:2006}. In this approach, we regress each variable
$X_j$ onto all other variables $(X_k)_{k\neq j}$ using the lasso. This
directly estimates the set of neighbors ${\mathcal N}(j) = \{k |
(j,k) \in E \}$ for each node $j$ in the graph, but the covariance
matrix is \textit{not} directly estimated. Lasso theory gives
conditions and guarantees on these variable selection problems. This
approach was adapted to the discrete case by \citet{Ravikumar:10}, where
the normalizing constant and thus the density can't be efficiently
computed. This general strategy may be attractive for graph selection
in nonparametric graphical models. In particular, each variable could
be regressed on the others using a nonparametric regression method that
performs variable selection; one such method with theoretical
guarantees is due to \citet{Rodeo}.

A different framework for nonparametricity involv\-es conditioning on a
collection of observed explanatory variables $Z$. \citet{Liu:gocart:10}
develop a nonparametric procedure called {\it Graph-optimized CART}, or
{\it Go-CART}, to estimate the graph conditionally under a Gaussian
model. The main idea is to build a tree partition on the $Z$ space just
as in CART (classification and regression trees), but to estimate a
graph at each leaf using the glasso. Oracle inequalities on risk
minimization and model selection consistency were established for
Go-CART by \citet{Liu:gocart:10}. When $Z$ is time, graph-valued
regression reduces to the time-varying graph estimation problem
(\citep{XiChen:10}; \citep{kolar09}; Zhou, Lafferty and Wasserman, \citeyear{Zhou:10}).

Another fruitful direction is the introduction of latent variables.
Even though the graphical model of the observed variables $X$ may be
complex, when conditioned on some latent explanatory variables $Z$, the
graph may be simplified. One straightforward approach is to build
mixtures of the models we consider here. A mixture of nonparanormals
will require new methods, to compute the derivatives $f'_j(x_j)$.
A~mixture of forests could be implemented using a kind of nonparametric
EM algorithm, with kernel density estimates over weighted data in the
M-step. But it is not easy to read off a graph from a mixture model.

In parametric settings, \citet{Chandrasekaran10} and \citet{Choi10}
develop algorithms and theory for learning graphical models with latent
variables. The first paper assumes the joint distribution of the
observed and latent variables is a Gaussian graphical model, and the
second paper assumes the joint distribution is discrete and factors
according to a forest. Since the nonparanormal and forest density
estimator are nonparametric versions of the Gaussian and forest
graphical models for discrete data, we expect similar techniques to
those of \citet{Chandrasekaran10}, \citet{Choi10} can be used to extend our
methods to handle latent variables. It would also be of interest to
formulate nonparametric extensions of low rank plus sparse covariance
matrices.

No matter how the methodology develops, nonparametric graphical models
will at best be approximations to the true distribution in many
applications. Yet, there is plenty of experience to show how incorrect
models can be useful. An ongoing challenge in nonparametric graphical
modeling will be to better understand how the structure can be
accurately estimated even when the model is wrong.

%
%

\end{document}